\pdfoutput=1
\documentclass[11pt]{article}

\usepackage[]{acl}

\usepackage{times}
\usepackage{latexsym}
\usepackage{amssymb}
\usepackage{pgfplots}
\usepackage{scalefnt}
\usepackage{boldline}
\usepackage{xcolor}
\usepackage{multirow}
\usepackage{makecell}
\usepackage{graphicx}
\usepackage{bm}
\usepackage{amsmath}
\usepackage{bbm}
\usepackage{float}
\usepackage{stfloats}
\usepackage{lipsum}
\usepackage{comment}
\usepackage{booktabs}
\usepackage{subfig}
\usepackage{lipsum}
\usepackage{cleveref}

\usepackage[flushleft]{threeparttable}

\usepackage{booktabs}
\usepackage[T1]{fontenc}
\usepackage[utf8]{inputenc}

\usepackage{microtype}

\usepackage{inconsolata}

\usepackage{tabularx}

\usepackage{xspace}

\usepackage{hyperref}
\usepackage{xcolor}
\usepackage{listings}
\usepackage{color}
\usepackage{xspace}
\usepackage{multirow}
\lstdefinestyle{data_prompt}{
    basicstyle=\rmfamily\scriptsize,
    breakatwhitespace=true,
    breaklines=true,
    showstringspaces=true
}
\lstdefinestyle{exp_prompt}{
    basicstyle=\rmfamily\footnotesize,
    breaklines=true,
    breakatwhitespace=true
}
\definecolor{purple1}{HTML}{7570b3}
\definecolor{green1}{HTML}{1b9e77}
\definecolor{orange}{HTML}{d95f02}
\definecolor{green2}{HTML}{4daf4a}
\definecolor{red}{HTML}{e41a1c}
\definecolor{blue}{HTML}{377eb8}
\definecolor{purple2}{HTML}{984ea3}

\newcommand{\ours}{\textsc{InstructIR}\xspace}

\usepackage{amssymb}% http://ctan.org/pkg/amssymb
\usepackage{pifont}% http://ctan.org/pkg/pifont

\title{\ours: A Benchmark for Instruction Following of Information Retrieval Models}

\author{
    Hanseok Oh{\textsuperscript{1}}{\thanks{\, \* Work performed during internship at LG AI.}} \quad 
    Hyunji Lee{\textsuperscript{1}} \quad 
    Seonghyeon Ye{\textsuperscript{1}} \quad 
    Haebin Shin{\textsuperscript{1}} \\ 
    \textbf{Hansol Jang{\textsuperscript{2}} \quad 
    Changwook Jun{\textsuperscript{2}} \quad 
    Minjoon Seo{\textsuperscript{1}}} \\
    {\textsuperscript{1}}KAIST AI\quad
    {\textsuperscript{2}}LG AI Research \\
    \texttt{\{hanseok, hyunji.amy.lee, vano1205, haebin.shin, minjoon\}@kaist.ac.kr} \\
    \texttt{\{hansol.jang, cwjun\}@lgresearch.ai} \\
}

\begin{document}
\maketitle

\begin{abstract}
    Despite the critical need to align search targets with users' intentions, retrievers often only prioritize query information without delving into the users' intended search context. 
Enhancing the capability of retrievers to understand the intentions and preferences of users, akin to language model instructions, has the potential to yield more aligned search targets.
% % Gap for improvementa / limitation
Prior studies restrict the application of instructions in information retrieval to a task description format, neglecting the broader context of diverse and evolving search scenarios. 
Furthermore, the prevailing benchmarks utilized for evaluation lack explicit tailoring to assess instruction-following ability, thereby hindering progress in this field.
% % Contribution 
In response to these limitations, we propose a novel benchmark, \ours, specifically designed to evaluate instruction-following ability in information retrieval tasks. 
Our approach focuses on user-aligned instructions tailored to each query instance, reflecting the diverse characteristics inherent in real-world search scenarios.
Through experimental analysis, we observe that some retrievers fine-tuned to follow task-style instructions, such as INSTRUCTOR~\citep{Su2022OneEA}, can underperform compared to their non-instruction-tuned counterparts. 
This underscores potential overfitting issues inherent in constructing retrievers trained on existing instruction-aware retrieval datasets~\footnote{Code and dataset are available at \url{https://github.com/kaistAI/InstructIR}}.
\end{abstract}

\section{Introduction}
% \section{Introduction}
% Motivation
    % retriever also needs to reflect instruction in Knowledge-Augmented LM scenario
    % Same query can be used with different Context / preference 
% GAP
    % retriever with instruction에 대한 task는 잘 정의되지 않았음
    % Task-side instruction을 반영하려는 움직임은 있었으나, query와 target 간의 관계가 이미 1:1인 evaluation set에서 instruction을 통해서 이해에 도움을 주는 것이었기에 instruction을 진짜 반영한 것인지 측정하기 어려웠음 
% What we want to do / expected contribution
    % instruction이 필수적인 evaluation set을 제시
    % 기존 retriever 모델들을 실험 & 분석 & finding 

% 1) Motivation 
    % # To help Knowledge-Augmented LM side needs
    % retriever should understand instructions too! 
% LM의 한계를 극복하기 위해서 retriever의 사용이 중요해짐
% LM side의 advance는 많았지만, Retriever는 end task에 맞게 튜닝되기 보다는 naive하게 쓰였음
% LLM에 들어가는 instruction 및 context에 대한 정보를 retriever도 함께 반영해서 타겟을 가져올 수 있다면 더 적합한 output이 생길 수 있을 것

Large Language Models (LLMs) are often further trained to align user instructions and preferences with instruction tuning for diverse generative tasks~\citep{Ouyang2022TrainingLM, wang2022self,zhang2023instruction}.
This kind of alignment to user preferences is also important for information retrievers to reflect diverse users' search intentions and preferences for the search targets.
For example, when a user writes a blog post for children about the current climate change issue, it may be better to search for articles that are easy to understand rather than complex scientific articles.
However, current retrievers often do not take this into account, focusing on utilizing only ambiguous queries even simplifying the details for users through reformulation~\citep{ma2023query}.
Moreover, lack of benchmarks to evaluate retrievers on user-aligned scenarios prevents the mature discussions of instruction following in retrieval task.

\begin{figure}[t!]
\begin{minipage}{\columnwidth}
\small
\centering
\includegraphics[width=1.0\columnwidth]{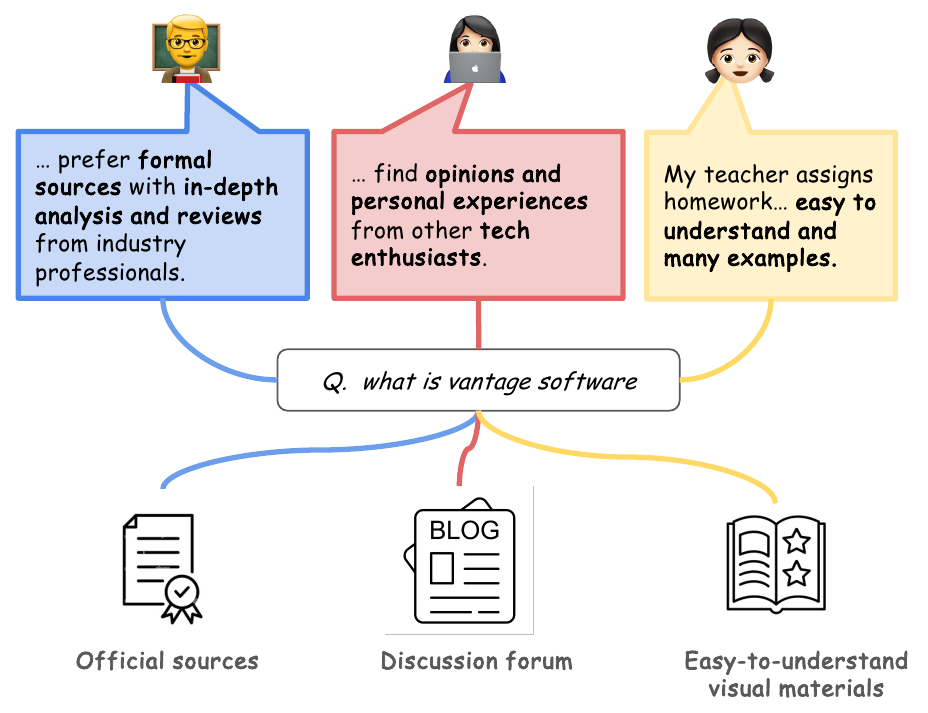}
\caption{\ours benchmark is designed to evaluate instruction following ability in information retrieval tasks. As unique user-aligned instructions change, different search targets should be retrieved to reflect real-world search scenarios.}
\label{fig: instructIR}
\end{minipage}
\vspace{-1em}
\end{figure}

\begin{figure*}[t!]
    \small
    \centering
    \includegraphics[width=0.8\textwidth]{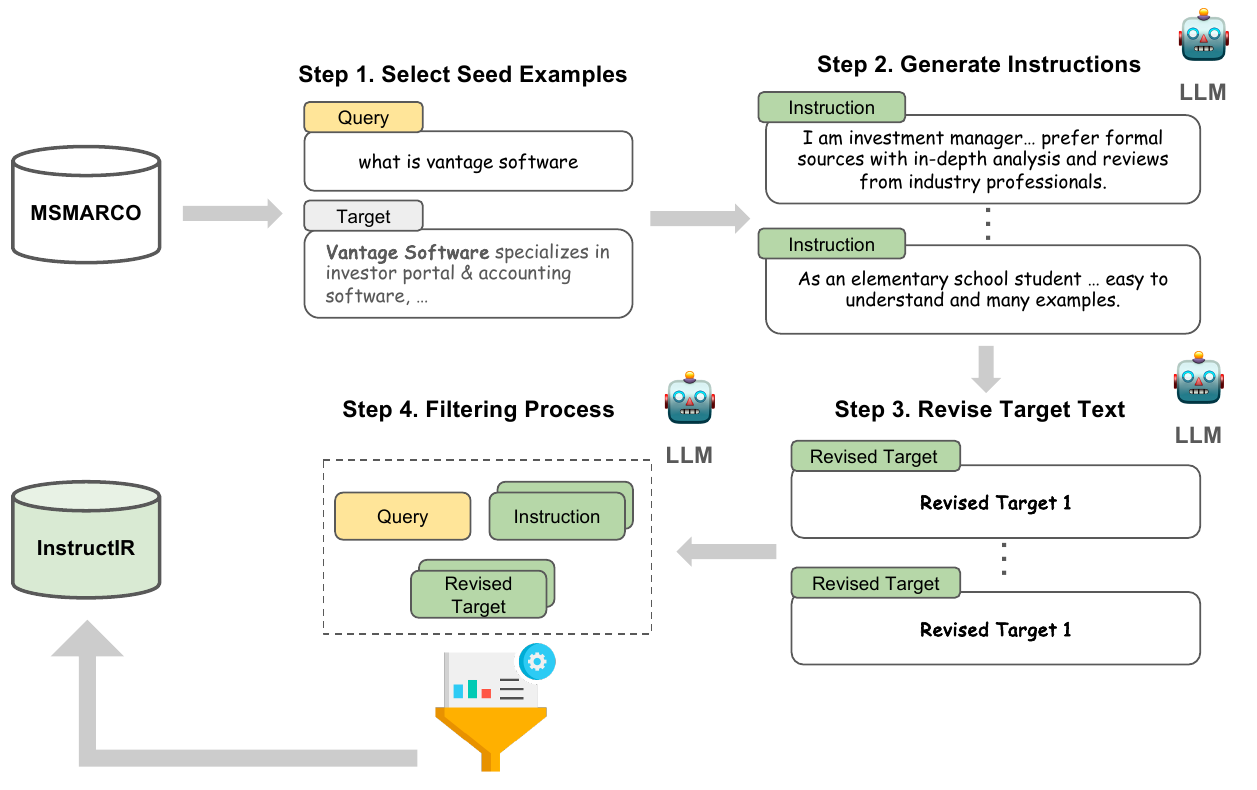}
    \caption{Overview of data creation pipeline for building \ours benchmark. To build datasets that demand diverse user-aligned instructions for each query, we begin by selecting seed examples from the MSMARCO datasets. Subsequently, we generate a variety of instructions suitable for each query, revise the target text to align with these instructions, and systematically filter the generated content. The resulting dataset is used for \ours benchmark. GPT-4 is employed in this generation pipeline.}
    \label{fig:data_creation_pipeline}
\end{figure*}

% 2) GAP
% retriever with instruction에 대한 task는 잘 정의되지 않았음
    % Task-side instruction을 반영하려는 움직임은 있었으나, query와 target 간의 관계가 이미 1:1인 evaluation set에서 instruction을 통해서 이해에 도움을 주는 것이었기에 instruction을 진짜 반영한 것인지 측정하기 어려웠음 

In order to effectively reflect the various intentions and situations that real-world users actually ask, employing instance-wise instructions for queries is more appropriate than relying on coarse-grained instructions that share the same task-specific guidance for various queries~\citep{wang2022self,ye2023flask}.
Several studies explore the integration of instructions into retrievers, but they primarily concentrate on building general purpose retrievers, which often limit the examination to task-specific instructions~\citep{asai-etal-2023-task,Su2022OneEA,Wei2023UniIRTA}.
For instance, these studies uniformly apply the same instructions to all instances, presenting them in the form of task descriptions, such as \textit{"Search for the Wikipedia paragraph that answers this question"}. 
Furthermore, the evaluation of these instruction-tuned models relies on benchmarks that inherently do not mandate instructions for task resolution~\citep{Thakur2021BEIRAH,Santhanam2021ColBERTv2EA,Muennighoff2022MTEBMT}.
Due to these limitations, the extent to which retrievers can effectively follow instructions has not been thoroughly evaluated.

In this work, we introduce a novel benchmark, \ours, specifically designed to evaluate instruction-following ability of retrieval models with diverse user-aligned instructions for each query, mirroring real-world search scenarios.
We collect a total of 9,906 of \textit{instance-wise instructions} that involve details about the search users, such as their job, background, situation, location, hobbies, additional interests, and search goals, and preferred sources.
Notably, \ours stands out from other benchmarks that evaluate task-aware instructions due to the distinctiveness in instruction types and diversity, as delineated in Table~\ref{tab:data_comparison_table}.
Instructions and corresponding search targets are acquired through our multi-stage data creation pipeline and filtering process, leveraging GPT-4~\citep{openai2023gpt4}, as illustrated in Figure~\ref{fig:data_creation_pipeline}.
The quality of the datasets is verified through a combination of human evaluation and machine filtering, resulting in a high-quality dataset. 
Additionally, we introduce the Robustness score as an evaluation metric, quantifying the ability of retrievers to robustly follow instructions. 
These metrics offer a holistic perspective on how effectively retrievers adapt to changes while conveying the same query with varying instructions.

% from not instruction-tuned models to instruction-tuned retrievers.
We evaluate over 12 retriever baselines on \ours including both na\"ive retrievers (not explicitly instruction-tuned) and instruction-tuned retrievers.
With our experiments, we find that task-style instruction-tuned retrievers, such as INSTRUCTOR~\citep{Su2022OneEA}, consistently underperform compared to their non-tuned counterparts, which cannot be found with the previous benchmarks.
% Furthermore, we  that employing a larger model leads to a meaningful enhancement in our benchmark results. 
Notably, utilizing an instruction-tuned language model and larger model as the backbone demonstrates the most potent performance improvement.
Through \ours, we gain valuable insights into the diverse characteristics of existing retrieval systems. 
We anticipate that this benchmark will contribute to accelerating progress in the development of more sophisticated, controllable, and instruction-aware information access systems.

\section{Related Works}
\paragraph{Evaluation for Instruction Following.} 
Insturction tuning is a crucial technique to enhance the capabilities and controllability of large language models (LLMs). 
Instruction tuning refers to the process of further training LLMs on a dataset consisting of \textit{(instruction, output)} pairs in a supervised fashion, which bridges the gap between the next-word prediction objective of LLMs and the users’ objective of having LLMs adhere to human instructions~\citep{Ouyang2022TrainingLM,Zhang2023InstructionTF}.
For the approaches to evaluate instruction following capabilities in generative tasks can be categorized as follows: Instructions for cross-task generalization, User Aligned instructions, and Verifiable instructions.
\textbf{Instructions for cross-task generalization} focus on evaluating cross-task generalization under instructions training models to follow instructions on a subset of tasks and evaluating them on the remaining unseen ones~\citep{wang2022super,liang2022holistic,wang2022super}.
\textbf{User Aligned instructions} focus on evaluating how instruction-based models handle diverse and unfamiliar instructions~\citep{ye2023flask, wang2022self, dubois2023alpacafarm}.
Unlike coarse-grained evaluation as for cross-task generalization instructions, they are different per instance. 
\textbf{Verifiable instructions} are focusing on straightforward and easy-to reproduce evaluation benchmark that focuses on a set of “verifiable instructions”~\citep{efrat2022lmentry,zhou2023instruction,jiang2023followbench}.

\paragraph{Instruction Following in Information Retrieval.} 
Building instruction-tuned models on text embedding tasks is mostly focused on building general purpose model that can solve multiple tasks with task description as an instructions~\citep{asai-etal-2023-task,Su2022OneEA,Wei2023UniIRTA}.
Models are trained with multiple source of train dataset with task descriptions as instructions, such as BERRI and MEDI, and evaluated on the held out tasks which haven't seen during the train time.
However, for solving these tasks it is not essential to follow instructions.
The relationship between query and targets are already in one to one relationship, where hard to evaluate effect of instructions. 
Moreover, the benchmarks used for evaluating instruction following ability are BEIR~\citep{Thakur2021BEIRAH}, LoTTE~\citep{Santhanam2021ColBERTv2EA}, X2~\citep{asai-etal-2023-task}, which are not suitable for evaluating the retrieval model’s ability to follow instructions because of coarse-grained evaluation per task not fine-grained instance-wise evaluation. 
MTEB~\citep{Muennighoff2022MTEBMT} is a similar type of benchmark that is widely used for evaluating instruction-tuned text embedding models, but it is not focused on retrieval tasks, so we do not compare it deeply.
Additionally, M-BEIR~\citep{Wei2023UniIRTA} is introduced for assessing multimodal retrieval tasks but is also constrained by a task-description style.

\section{The \ours Benchmark}
\subsection{Data Creation Pipeline}
Constructing a framework to evaluate instruction-following capabilities in information retrieval models necessitates correlating multiple instructions with the same query and adjusting their targets accordingly (i.e., instruction, query, target text).
Therefore, in contrast to previous approaches that evaluate coarse-grained task description-style instructions on information retrieval datasets with up to 15 instructions, we focus on creating per-query, instance-specific instructions as Table~\ref{tab:data_comparison_table}.
We employ GPT-4~\footnote{We use \texttt{gpt-4-1106-preview} for our work.}~\citep{openai2023gpt4} to facilitate the creation of such a setting.
The development of our \ours evaluation datasets adheres to a systematic set of steps as Figure~\ref{fig:data_creation_pipeline}, outlined as follows:

\paragraph{Step 1. Select Seed Examples.}
In tackling the challenge of generating all components from scratch, we opt to levarage the MSMARCO dataset~\citep{nguyen2016ms} for passage ranking into our seed examples~\footnote{We utilize the validation split from MSMARCO, consisting of 6,980 queries.}.
This dataset is renowned for its comprehensive coverage of diverse topics collected from the real web.
We carefully select a total of 1,743 queries $Q$ and corresponding target texts $T$ as seed examples, adhering to the following criteria:
1) The seed query should demonstrate the potential to match various targets as instructions evolve.
2) The target should offer substantial content, allowing for modifications that align with provided instructions.
3) Ensuring ease in controlling false negatives is crucial.
To meet the first criterion, we focus on queries with a length ranging from 25\% to 75\% (i.e., 24 - 40). 
This approach prevents the seed query, pivotal for formulating subsequent instructions, from being overly vague or verbose. 
Similarly, for the second criterion, aligning with the rationale for query selection, we chose target texts within the text length range of 255 - 371 to avoid ambiguity. 
Lastly, for the third criterion, to facilitate the control of potential false negatives, we exclusively extract instances with only one positive target.

\paragraph{Step 2. Generate Instructions.}
Following the careful selection of seed examples in the previous step, we harness the power of GPT-4 to produce a set of instructions $I_i$ corresponding to each query $q_i$, where $i$ ranges from 1 to $n$, denoting the number of queries.
To evaluate the adherence of retrievers to instructions, we emphasize the importance of deploying the same query with different instructions as illustrated in Figure~\ref{fig: instructIR}. 
This approach enables us to measure how effectively models dynamically retrieve relevant targets.
We also insist to move away from defining instructions in the form of rigid task descriptions, instead to embrace a more realistic approach that captures real-world scenarios in which retrieval systems are employed. 
This involves incorporating diverse information about users, such as their occupation, search context, location, search objectives, and preferred sources. 
To achieve this, we adopt the prompt outlined in Figure~\ref{fig:data_generation_step2_prompt}. 
We produce a set of instructions $I_i = \{I_{i,1}, ..., I_{i,k}\}$ (where $k$ is set to 10) for each query $q_i$ with a particular focus on aligning these instructions with a precise reflection of scenarios that real users may encounter.

\paragraph{Step 3. Revise Target Text.}
% prompt used / need example
During this phase, GPT-4 refines the original target text $T$ by integrating the instructions $I_i$ generated in step 2.
The model takes in a query $q_i$, an instruction $I_{i,k}$, and the original target $t_i$ derived from seed examples.
% Subsequently, GPT-4 adjusts the target to better align with the provided instructions, resulting in $t_i^{'}$. 
Subsequently, GPT-4 adjusts the target to better align with the provided instructions, resulting in $t_{i,k}^{'}$. 
This adjustment involves revising the target to accurately represent the given scenario, taking into account factors like the user's background, situation, location, occupation, hobbies, interests, or goals for the search. 
Additionally, it is crucial to incorporate information related to the user's preferences.
To ensure the generation of diverse targets in response to evolving instructions, we employ a prompt, illustrated in Figure~\ref{fig:data_generation_step3_prompt}.

\paragraph{Step 4. Filtering Process.}
To assess the quality of machine-generated datasets during steps 2 and 3, we proceed by filtering out datasets in this stage.
The selection of high-quality instances is based on the evaluation of two key criteria: \textit{Q1. Does the revised target align with the original query?} and \textit{Q2. Does the revised target align with the given query and instructions, while other targets do not?}. 
Additionally, we leverage the capabilities of GPT-4 as a quality evaluator.
% For Q1, we employ the prompt presented in Figure~\ref{fig:filtering_c1_prompt} to retain instances with a high score between the original query $q_i$ and the revised target text $t_i^{'}$~\footnote{Empirically, we observe that a score exceeding 3 out of 5 indicates good quality.}.
For Q1, we employ the prompt presented in Figure~\ref{fig:filtering_c1_prompt} to retain instances with a high score between the original query $q_i$ and the revised target text $t_{i,k}^{'}$~\footnote{Empirically, we observe that a score exceeding 3 out of 5 indicates good quality.}.
% To address Q2, the prompt in Figure~\ref{fig:filtering_c23_prompt} is utilized to identify instances where GPT-4 accurately predicts the gold target $t_i{'}$ among distractors $t_j^{'}$ (where $j\ne i$), generated from the same query with different instructions. 
To address Q2, the prompt in Figure~\ref{fig:filtering_c23_prompt} is utilized to identify instances where GPT-4 accurately predicts the gold target $t_{i,k}^{'}$ among distractors $t_{i,m}^{'}$ (where $m\ne k$), generated from the same query with different instructions. 
% This is based on a high score between original query $q_i$ and the revised target text $t_i^{'}$. 
This is based on the scores between the original query $q_i$ and an instruction $I_{i,k}$ pairs, and the set of revised target text $T_i'$. 
Following the filtering stage, we select instances that possess more than 6 instructions with the same query, facilitating an effective evaluation of retrievers in dynamically changing scenarios. 
This results in a total of 9,906 instances, as detailed in Table~\ref{tab:data_statistics}. 
Further statistics regarding the filtering stage and examples can be found in the Appendix~\ref{appendix:dataset_details}.

\subsection{Dataset Analysis}

\begin{table}[t!]
\centering
\resizebox{1.0\columnwidth}{!}{
\begin{tabular}{cccccc}
\toprule
   & User-aligned & Type of inst.& \# of inst. & Metrics\\
  \midrule
\ours & \ding{51} & instance-wise & 9,906 & Robustness, nDCG\\
\midrule
BEIR  & \ding{55} & task-wise & 15  & nDCG\\
\midrule
LoTTE & \ding{55} & task-wise & 5  & nDCG \\
\midrule
X2 & \ding{55} & task-wise & 6  & nDCG \\
\bottomrule
\end{tabular}%
}
\caption{Table comparing \ours with other IR benchmarks used to measure instruction following ability. 
Since the other benchmarks are not designed to evaluate instruction-following ability of information retrieval task, they are based on reformulated versions from previous studies~\citep{asai-etal-2023-task,Su2022OneEA}.
}
\label{tab:data_comparison_table}
\end{table}
% half column으로 바꾸기?
% User aligned instruction은 shorten
% # of instruction => shorten 
% 어디서 해당 benchmark를 썼는지는 caption정도에 넣어주기
% MTEB, MBEIR 는 related works정도에 언급만

\paragraph{Comparison Table.}
We characterize our \ours dataset as shown in Table~\ref{tab:data_comparison_table}. 
Note that all other benchmarks (BEIR, LoTTE, and X2) are not proposed for evaluating instruction following ability of retrieval systems originally, it is reformulated with author proposed instructions in the previous studies \citep{asai-etal-2023-task,Su2022OneEA,wang2023improving}.
Other benchmarks are not focused on user aligned scenario, leading to structured formats that are far from search scenario of real users.
Also, others utilize task-specific instructions, where up to 15 instructions are used for the evaluation, which is too small to evaluate whether a retrieval model can follow the instructions.

% \paragraph{Diversity and Data Quality.}
\paragraph{Dataset Quality.}
To ensure the quality and authenticity of our generated and filtered datasets, a human verification stage is implemented for validation purposes.
For about 8\% of the randomly sampled groups, a total of 10 annotators evaluate two instances per group.
For each instance, we pose three questions, and three annotators are assigned to assess them, aiming to measure the inter-agreement between annotators~\citep{Cohen1960ACO, Randolph2005FreeMarginalMK}.
We consider the final human decision by majority voting.
Further detail settings and results with inter-agreement are available in Appendix~\ref{appendix:human_verification_details}.
The first question (\textit{Q1: Is the instruction valid for the search user?}) asks whether the provided instruction is suitable for the search user scenario. For this question, all instances were answered suitable for the search scenario.
The second question (\textit{Q2: Is the instruction natural for the given query?}) asks whether the instruction is naturally aligned with the query. About 97\% instances were answered that the instruction and query have well-aligned relations.
The third question (\textit{Q3: Which passage is the most natural for the given instruction and query?}) asks to choose the most relevant passage. This question can estimate the difficulty of this benchmark, and evaluate the alignment between the results of Q2 from the filtering process (step 4) and human judgment. As a result, we got kappa coefficient ~\citep{Cohen1960ACO} of 0.6468 which indicates substantial agreement between human judgement and dataset~\citep{Landis1977TheMO}. 
However, annotators demonstrate an approximately 76.5\% top-1 accuracy, emphasizing the need for careful consideration when selecting the top-1 passage that follows the instructions.
User interface and instruction for annotators are described in Figure~\ref{fig:human_verification_user_interface}.
% As in Table~\ref{tab:human_verification} ~. 

\begin{table}[t!]
\centering
\resizebox{1.0\columnwidth}{!}{
\begin{tabular}{ll}
\toprule
   & Number \\
\midrule
Avg. instructions per query & 7.81 \\
number of queries & 1,267 \\
Number of query with instructions & 9,906 \\
number of corpus & 16,072 \\
relevancy & binary \\
\bottomrule
\end{tabular}
}
\caption{Statitistics for InstrucIR dataset}
\label{tab:data_statistics}
\vspace{-1em}
\end{table}

\paragraph{Dataset Diversity and Statistics.}
Table~\ref{tab:data_statistics} presents the data statistics for datasets within \ours. 
Through our data creation pipeline, we acquire an average of 7.81 instructions per query. 
Furthermore, to simulate real-world noisy search scenarios, we incorporate a subset of targets from the seed datasets. 
This results in an augmentation of approximately 6k additional targets, supplementing our revised targets.
Consequently, our benchmark comprises a total of 10k instances, featuring a rich variety of instructions and a corpus with 16k entries, collectively constituting the \ours benchmark. 
Recognizing the significance of encompassing diverse instructions for the same query to reflect various user search scenarios, we evaluate the diversity of our dataset using the ROUGE score distribution inspired by \citet{wang2022self}. 
Specifically, we calculate the average ROUGE-L score for instructions associated with the same query.
As illustrated in Figure~\ref{fig:rouge}, the instructions within \ours encapsulate highly diverse scenarios with low similarity to each other, achieving an average ROUGE-L score of 0.238 within each group.

\subsection{Evaluation Metric}
We employ the Normalized Cumulative Discount Gain (nDCG@k) following \citet{Thakur2021BEIRAH}, as it is a widely accepted evaluation metric for assessing retrieval models.
However, we emphasize the need for specialized metrics to assess the ability to follow instructions when retrieving information, especially when measuring dynamic changes in retrieval targets as instructions for a given query change.
In instances where diverse search scenarios and user preferences serve as instructions alongside a given query, retrieval systems should adapt to these instructions to identify appropriate targets effectively. 
Hence, inspired by \citet{zhong2022romqa} and \citet{oh2023ktrl+}, we introduce a Robustness score to assess how consistently the model predict targets over evolving instructions using the same query.
To quantify robustness, we group instances with identical queries, calculate the minimum nDCG@k score within each group, and subsequently average these per-group scores to derive the final Robustness@k score.

\section{Experiments}
\begin{table*}[t!]
\resizebox{\textwidth}{!}{%
\begin{tabular}{cccccc}
\toprule
Instruction-tuned & Type & Models & Size & Robustness@10 & nDCG@10 \\
\midrule
\multirow{6}{*}{No} &  Lexical & BM25 & - & 26.92 & 76.01 \\
    & Late-Interaction & ColBERT-v2.0 & 110M & 14.15 & 68.47 \\
   & \multirow{5}{*}{Bi-Enc.} 
    &  Contriever-msmarco & 110M & 47.40 & 84.85 \\
    & & GTR-base & 110M & 34.06 & 73.35 \\
    & & GTR-large & 335M & 37.56 & 75.95 \\
    & & GTR-XL & 1.5B & 38.34 & 75.20 \\
    & & RepLLaMa & 7B   & 52.58 & \textbf{87.62} \\
    \midrule
\multirow{4}{*}{Yes} & \multirow{4}{*}{Bi-Enc.} 
    &TART-dual & 110M & 47.46 & 84.81\\
    & & INSTRUCTOR-base & 110M & 23.73 & 50.44 \\
    & &INSTRUCTOR-large & 335M & 22.08 & 48.80 \\
    & & INSTRUCTOR-XL & 1.5B & 21.53 & 48.63 \\
    & &E5-mistral-7b-instruct & 7B & \textbf{55.42} & 86.33\\
\bottomrule
\end{tabular}%
}
\caption{Zero-shot performances on \ours benchmark. We group models based on whether instruction-tuned or not, type of scoring methods, and the size of the models. Bi-Encoder is abbreviated as Bi-Enc.}
\label{tab:main_table}
\end{table*}

\subsection{Baselines}
We utilize \ours to compare various retriever systems in a zero-shot manner.
\ours exclusively offers test datasets designed to evaluate the proficiency of existing retrievers in addressing information retrieval tasks that require adherence to specific instructions.
In our experimentation, we use pre-trained checkpoints accessible online for all models. 
We categorize the models based on the following criteria: non-instruction-tuned retrievers and instruction-tuned retrievers.

\paragraph{Non-instruction-tuned Retrievers.}
For non-instruction-tuned retrievers, we select BM25~\citep{robertson2009probabilistic}, Contriever-MSMARCO~\citep{Izacard2021UnsupervisedDI}, GTR~\citep{ni2022large}, and RepLLaMa~\citep{ma2023fine}.
BM25 represents a lexical matching retriever. 
Contriever-MSMARCO, a bert-base sized model, is fine-tuned on the MSMARCO~\citep{nguyen2016ms} passage ranking dataset following an unsupervised contrastive pre-training stage. 
GTR comprises variants of a t5-encoder based bi-encoder, trained on both MSMARCO and NQ datasets. 
RepLLaMa is a bi-encoder model based on the LLaMa-2-7b decoder, extracting token embeddings from the end-of-sequence token to generate text embeddings.

\paragraph{Instruction-tuned Retrievers.}
For instruction-tuned retrievers, we utilize TART-dual~\citep{asai-etal-2023-task}, INSTRUCTOR~\citep{Su2022OneEA}, and E5-mistral-7b-instruct~\citep{wang2023improving}.
TART-dual is a retriever with a bi-encoder architecture based on the Contriever model.
It is additionally trained on BERRI, an instruction-aware information retrieval dataset comprising approximately 40 diverse retrieval tasks, each accompanied by a task description serving as instructions.
INSTRUCTOR offers various size versions, including base, large, and xl, all finetuned on the GTR retriever as a backbone~\footnote{As INSTRUCTOR also leverages corpus-side instruction, we adhere to its approach by using the corpus instruction: \textit{'Represent the document for retrieval:'} utilized for the MSMARCO dataset.}. 
It is further trained on the MEDI dataset, encompassing 330 distinct text embedding tasks, each with a human-written task instruction, including multiple retrieval datasets.
Lastly, E5-mistral-7b-instruct is a bi-encoder architecture retriever trained on a proprietary Language Model(LLM), the mistral-7b decoder. 
It undergoes training solely with synthetic data for text embedding tasks, with task definitions serving as instructions.

% \paragraph{Oracle.}
% % Using GPT-4 as reranker 
% To assess the potential upper bound of performance for \ours, we conduct additional experiments using a Large Language Model(LLM) as a cross-encoder architecture.
% Given the complexity of ranking all pairs from the corpus with a cross-encoder design, we mitigate this challenge by narrowing down the candidates for ranking based on gold documents that share the same query but have different instructions.
% For this purpose, we employ the open-sourced language model Vicuna-13b as a cross-encoder. 
% The ranking of candidate documents for a given instance is conducted using a list-wise approach, a methodology frequently adopted in diverse prior studies that utilize Language Models (LLM) as rerankers~\citep{ma2023zero, sun2023chatgpt}.

\subsection{Results}
% Table~\ref{tab:main_table} shows performance of diverse retrieval baselines in \ours. 
% Among all, E5-mistral-7b-instruct shows most powerful result in both nDCG@10 and Robustness@10. 
% And models with high lexical 
We evaluate various retrieval systems on \ours benchmark to evaluate their capability to follow instructions in a zero-shot setting as Table~\ref{tab:main_table}. 
The baselines are categorized into instruction-tuned and non-instruction-tuned models across different sizes and architectures.
\paragraph{Non-Instruction-Tuned Models.}
Considering overall high score of nDCG over 50, due to the characteristics of distinct instructions in \ours datasets, lexical bias exists.
However, the lexical matching model BM25 and late interaction model ColBERT-v2.0 show Robustness@10 of 26.92 and 14.15 respectively.
And it means that focusing individual keywords can't truly understand evolving instructions, which leads to huge drop in Robustness score.
Representative retrievers that are trained with contrastive training objectives, such as Contriever-msmarco and GTR variants (base, large, and xl), are considered, with sizes ranging from 110M to 1.5B parameters. 
These models exhibited a significant improvement in Robustness over BM25, which Contriever-msmarco shows most powerful and robust performance compared to the same size models in non-instruction tuned baselines even superior to way more larger size GTR-xl. 
The largest model, RepLLaMa with 7B parameters, achieve the highest nDCG@10 of 87.62 and Robustness@10 of 52.58, indicating a strong correlation between model size and performance metrics in non-instruction-tuned settings.

\paragraph{Instruction-Tuned Models.}
TART-dual undergoes additional training steps with BERRI, a comprehensive set of over 40 retrieval datasets accompanied by task-specific instructions generated by human experts, based on Contriever-msmarco. 
INSTRUCTOR variants are subsequently trained, aligning with the respective sizes of GTR retrievers, using MEDI datasets. 
These datasets are reformulated training sets covering 330 datasets, incorporating instructions spanning diverse text embedding task categories and domains, including retrieval tasks.
However, both series of instruction-tuned baselines do not show superior performance than their non-instruction-tuned counterparts. 
Notably, INSTRUCTOR variants exhibit a huge performance drop compared to the backbone model GTR variants. 
This can be interpreted that finetuning retrievers with only task-style instructions doesn't guarantee good performance in various free creative user-aligned style instructions. 
Conversely, E5-mistral-7b-instruct, shows Robustness@10 with 55.42 outperforming all other models.
This highlights the importance of using large instruction-tuned models for search tasks to follow instructions as well.

\section{Analysis}
% 결과
% - instruction tuned vs. non-instruction tuned
% 분석
% - Scale
%     - GTR 은 오히려 잘된다
%     - INSTRUCTOR 오히려 클수록 overfitting이 잘돼서 반대로 떨어진다?
%     - repllama, mistral
% - 순서 민감성
% - prompt sensitivity
%     - instruction paraphrase
% - lexical overlap
%     - bm25
% - Colbert
%     - MaxSIM ?

\paragraph{Scaling Up Model Size Leads Better Instruction Following.}
It is not surprising that larger models derive greater benefits from instruction tuning, indicating that their enhanced capacity enables a more effective integration of instruction-following abilities. 
In the case of non-instruction-tuned baselines, GTR demonstrates superior performance as model sizes increase, particularly in terms of Robustness scores. 
Remarkably, Contriever-msmarco exhibits competitive performance even with smaller sizes.
However, INSTRUCTOR variants display reverse trends, resulting in lower Robustness scores as model sizes increase. 
This hints at an overfitting issue to diverse task description style instructions, leading to diminished performance across varying instruction styles and longer, unseen user-aligned scenarios.
Nonetheless, the outstanding performances of 7B size models, RepLLaMA, and E5-mitral-7b-instruct underscore the significance of both model size and instruction tuning in developing competent retrieval systems for complex, instruction-based queries.

\paragraph{Instruction Order Sensitivity Exists For Instruction Tuned Retrievers.}
To analyze the importance of the order of instructions and queries, we conduct additional experiments by changing their sequence (query-> instruction). 
Remarkably, INSTRUCTOR exhibits a significant performance gain when query precede the instructions as Table~\ref{tab:analysis-instruction-order}.
Considering the average length of instructions in the training dataset used for INSTRUCTOR, it is approximately 12.16 tokens. 
In contrast, the average number of tokens for instructions utilized in our user-aligned instructions is about 64.47. 
This discrepancy highlights the challenge of generalizing models trained solely on task-description style instructions, which exhibit limited creativity and variety, to more user-aligned cases.
For a detailed comparison between coarse-grained task description instructions and user-aligned instance-specific instructions, please refer to the examples provided in the Appendix~\ref{appendix:dataset_example}.

\begin{table}[]
\resizebox{\columnwidth}{!}{%
\begin{tabular}{lll}
\toprule
Models                 & Robustness Gap & nDCG Gap \\
\midrule
BM25                   & -17.62 &  -14.69 \\
Contriever-msmarco     & -2.31 & -1.78\\
ColBERT-v2.0           &+22.20 & +10.45\\
GTR-base               &-2.62 & -1.53\\
GTR-large              &-1.97  & -1.34 \\
GTR-XL                 &-2.53  & -1.22 \\
RepLLaMa               & -4.11 & -2.94\\
TART-dual              &-0.58  &  -0.82\\
INSTRUCTOR-base        & + 17.06 &  + 32.82\\
INSTRUCTOR-large       & \textbf{+ 34.70} &  \textbf{+ 40.74}\\
INSTRUCTOR-XL          & +23.26 & + 37.27\\
E5-mistral-7b-instruct & +2.89 & +1.61\\
\bottomrule
\end{tabular}%
}
\caption{Performance gap for changing instruction order.}
\label{tab:analysis-instruction-order}
\vspace{-1em}
\end{table}

\begin{figure}[t!]
\begin{minipage}{\columnwidth}
\small
\centering
\includegraphics[width=1.0\columnwidth]{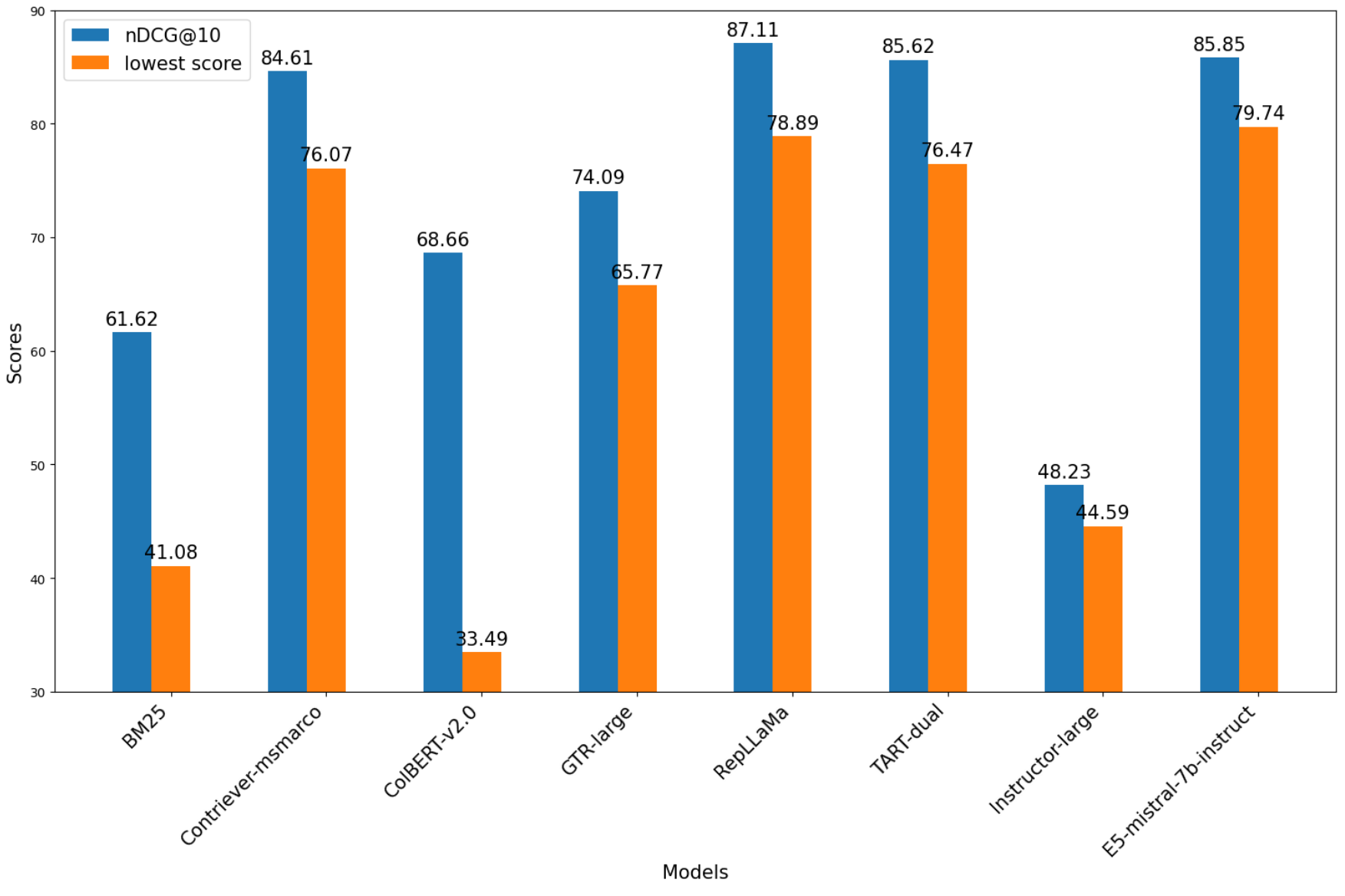}
\caption{Prompt sensitivity per models. Blue bar and orange bar denote performance of original instructions and smallest score of paraphrased versions respectively.}
\label{fig:analysis-prompt-sensitivity}
\end{minipage}
\end{figure}

\paragraph{Weighting Individual Terms leads Instruction Sensitivity for the Paraphrased Instructions.}
To analyze the sensitivity of instructions when paraphrased, we randomly select one instance from each group in \ours, resulting in 1,267 subsets of instances. 
Subsequently, we generate five paraphrased versions for each instance using GPT-4. 
This process yields a total of 1,267 groups of paraphrased instructions. 
The evaluation is based on the smallest score among the five paraphrased instructions.
As illustrated in Figure~\ref{fig:analysis-prompt-sensitivity}, retrievers specialized in emphasizing specific terms, such as BM25 and ColBERT-v2.0, exhibit a substantial performance drop of 20.53 and 35.17, respectively. 
In contrast, most bi-encoder based models demonstrate less fluctuation compared to these two models. 
Notably, the E5-mistral-7b-instruct model displays the most robust performance in adapting to changing instructions.

% \paragraph{Lexical Overalp.}
\paragraph{Relying on Lexical Redundancy Reduces Robustness.}

\begin{table}[]
\resizebox{\columnwidth}{!}{%
\begin{tabular}{lll}
\toprule
Models                 & Robustness Gap & nDCG Gap \\
\midrule
% BM25                   & -26.92             & -76.01 \\
Contriever-msmarco     & +4.10             & -25.31\\
ColBERT-v2.0           & \textbf{+10.91} & \textbf{-36.43}\\
GTR-base               & +3.09              & -27.95 \\
GTR-large              & +5.95              & -24.48 \\
GTR-XL                 & +4.65              & -24.55\\
RepLLaMa               & +1.35               & -24.96\\
TART-dual              & +6.57              & -22.93 \\
INSTRUCTOR-base        & +6.65              & -14.73 \\
INSTRUCTOR-large       & +9.62              & -11.80 \\
INSTRUCTOR-XL          & +10.74             & -11.36 \\
E5-mistral-7b-instruct & +6.90              & -17.14\\
\bottomrule
\end{tabular}%
}
\caption{Performance gap when removing high lexical overlap. BM25 is removed as used for filtering.}
\label{tab:analysis-lexical-overlap}
\end{table}

Considering existence of distinct information per the instruction, lexical hint can exist to find proper target.
% Thus, we remove high lexical overlap case by filtering out instances that BM25 predicts ground truth answer in top 10, and how other semantic matching models change. 
Therefore, we eliminate instances with high lexical overlap by filtering out cases where BM25 predicts the ground truth answer in the top 10, and evaluate how other semantic matching models are affected by this filtering process.
Overall, average Robustness score for the less lexical overlap cases increases in all model, and nDCG score drops.
And it means some of datasets in \ours can be solved by lexical matching, but relying heavily on lexical cues can lead to wrong targets, which leads to lower Robustness score.
Among all, ColBERT-v2.0 shows the largest changes showing average plus 10.91 Robustness score, and minus 36.43 nDCG score. 
RepLLaMA shows least Robustness performance change.

% \paragraph{Effect of MaxSim Operations for Late-Interaction Retrievers.}

\section{Conclusion and Future Work}
In this study we introduce a novel benchmark, \ours, designed for evaluating the instruction following capabilities of information retrieval models. 
Despite the critical importance of aligning models with user instructions and reflecting user preferences in information retrieval tasks, existing evaluations often fall short in comprehensively assessing these aspects. 
Therefore, our benchmark focuses on evaluating user-aligned instructions tailored to each query instance, reflecting the diverse characteristics inherent in real-world search scenarios.
Our experimental investigation sheds light on the instruction following capabilities of information retrieval models, presenting valuable insights that contribute to the current understanding of this domain. 
In particular, we highlight the gaps in current instruction-tuned retrievers in the limited style of training datasets organized around task description style instructions.

One promising direction for future research is the exploration of methodologies, such as Reinforcement Learning from Human Feedback (RLHF), to enhance the alignment of retrieval models with users' search intentions as proposed by \citet{Ouyang2022TrainingLM}. 
Investigating the integration of RLHF techniques can potentially lead to more effective and adaptive information retrieval systems that better understand and respond to user instructions. 
Additionally, future studies could delve into the development of more diverse instruction-aware retrieval training datasets that capture the nuances of user preferences and instructions in a more intricate manner. 
By addressing these challenges, we anticipate significant advancements in the field, ultimately improving the overall user experience in information retrieval scenarios.

\section{Limitations}
% However, it is imperative to acknowledge certain limitations inherent in our approach. 
% Firstly, our proposed benchmark, \ours, may possess specific characteristics that limit its ability to capture the full spectrum of user instructions or preferences, potentially hindering its generalizability.
% A challenge lies in accurately simulating real-world user instructions within an automatic data creation setting.
% While our study incorporates user-aligned information related to job, background, situation, location, occupation, hobbies, interests, goals of the search, and document preferences, replicating the complexity of actual user interactions remains a formidable task. 
% The creation of a more comprehensive set of instructions, spanning different domains and incorporating diverse seed datasets, could provide a more nuanced evaluation.
% Moreover, future research endeavors could extend the experimental focus beyond the zero-shot evaluation of existing instruction-tuned and non-instruction-tuned retrievers. Exploring the potential for performance improvement through proper instruction-aware training would enhance our understanding of the models' adaptability to user instructions.
% Addressing these limitations will be pivotal for advancing our comprehension of instruction following in information retrieval models. 
% It is crucial to develop more robust and universally applicable evaluation frameworks that consider diverse user inputs and domain-specific nuances, ultimately improving the overall effectiveness and user satisfaction in information retrieval scenarios.

While our proposed benchmark, \ours, leverages a representative retrieval dataset that mirrors real user behavior patterns and has undergone systematic filtering processes and human validation steps, the challenge lies in accurately simulating real user instructions within an automated data generation setting.
Our study incorporates user-aligned information related to job, background, situation, location, occupation, hobbies, interests, goals of the search, and document preferences. 
However, there is room for improvement by enhancing the comprehensiveness of instructions through the integration of diverse seed datasets across various domains, thereby offering a more comprehensive evaluation.
Additionally, we conduct experiments in a zero-shot setting, where the retrievers are not explicitly tuned based on user-aligned instructions, aiming to assess the adaptability of current instruction-tuned retrieval systems to different styles of instructions.
Further analysis to explore the potential for performance improvement through user-aligned instruction-aware training would deepen our understanding of the models' adaptability to user instructions in the future.

% \section*{Acknowledgements}

\bibliographystyle{acl_natbib}
\bibliography{anthology,custom}

\clearpage
\newpage
\appendix

\section*{Appendix}
\label{sec:appendix}
\section{Details for Dataset Construction Pipeline}
\label{appendix:dataset_details}

\begin{figure*}[p]
\begin{lstlisting}[style=data_prompt]
>> SYSTEM PROMPT
You are a helpful, respectful and creative assistant.

>> USER INSTRUCTION
Your task is to generate a set of scenarios for the provided search query.
Here is the specification for the scenario generation task: 
- The scenario should reflect a very specific scenario where a user is interacting with an AI search engine. 
- Within the scenario, the user could write about his/her job, background, situation, location, occupation, hobbies, interests, or goals of doing the search. Also, the user could explicitly reflect about his/her preference regarding the document to be searched.
- The scenario SHOULD be written from a first person's view point. For example, it should start with phrases like "I am a {job}, ...", "I am in a situation ...", "During my {situation}".
- While the provided query is about "what" is being searched for, the scenario you will generate should be about "how" the search should be approached and what values or criteria should be prioritized in that search. This distinction makes the role of the scenario clear as a guiding framework for the search process, as opposed to the query which is more about the specific target of the search.
- However, the scenario should be RELATED with the provided query. In other words, it shouldn't be applicable to other queries in general.
- You should generate based on the following format (note that there is a phrase 
"[END]" after each scenario being generated):
Scenario 1: {generate the first scenario} [END]
Scenario 2: {generate the second scenario} [END]

Scenario 10: {generate the last scenario} [END]
- Please do not generate any other opening, closing, and explanations. Just generate the set of scenarios!
\end{lstlisting}
\caption{Prompt for generating instructions (step 2)}
\label{fig:data_generation_step2_prompt}
\end{figure*}

\begin{figure*}[p]
\begin{lstlisting}[style=data_prompt]
>> System Prompt 
You are a helpful, respectful and creative assistant.

>> USER INSTRUCTION
Your task is to generate a REVISED DOCUMENT for the provided search QUERY and SCENARIO pair.
Here is the specification for the DOCUMENT revising task:
- The REVISED DOCUMENT should reflect the user's unique SCENARIO where a user is interacting with an AI search engine.
- Within the REVISED DOCUMENT, revise details reflecting the user's background, situation, location, occupation, hobbies, interests, or goals of doing the search. Also, containing information related to the user's preference is important. 
- Directly revise given DOCUMENT that has good quality that can be found by an AI search engine. Don't just suggest it!
- Do NOT include the same keywords from the given SCENARIO in REVISED DOCUMENT. Paraphrase it. 
- However, the REVISED DOCUMENT should be RELATED with the provided query. In other words, it should be applicable to query in general.
- You should generate based on the following format (note that there is a phrase "[END]" after each elements being generated):

PLAN: {generate the plan for the strategy for revision} [END]
REVISED DOCUMENT: {revise the document} [END]

- Please do not generate any other opening, closing, and explanations. Just generate the PLAN and REVISED TARGET !
\end{lstlisting}
\caption{Prompt for revising target text (step 3)}
\label{fig:data_generation_step3_prompt}
\end{figure*}

\begin{figure*}[t!]
\begin{minipage}{\textwidth}
\small
\centering
\includegraphics[width=0.8\columnwidth]{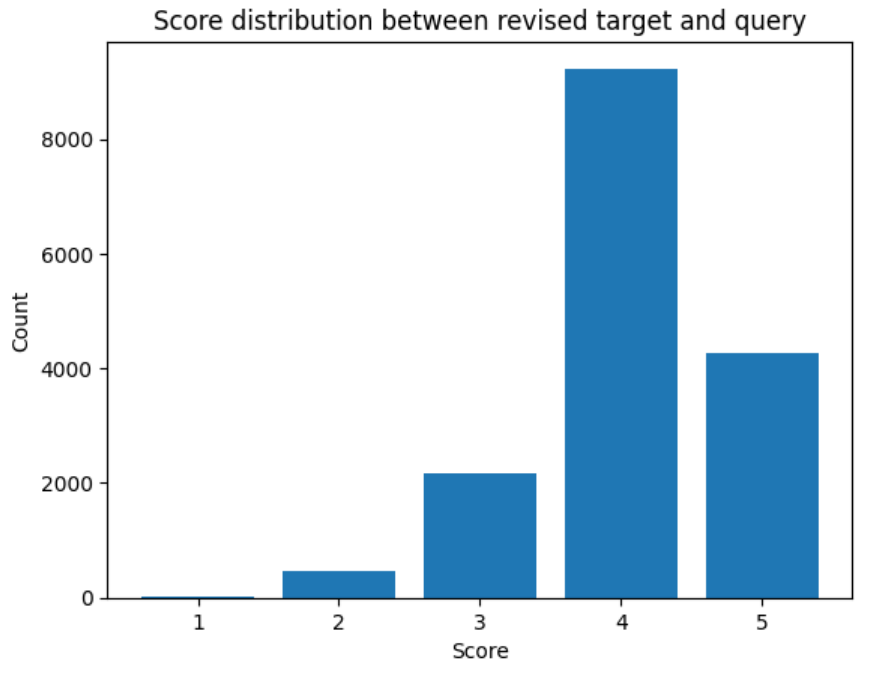}
\caption{Relevancy score distribution between revised target after data generation step3 and query.}
\label{fig:filtering_c1}
\end{minipage}
\end{figure*}

\paragraph{Filtering Process.}
During the filtering process in step 4 of data creation pipeline, we perform two filtering criteria.
For the first criterion, \textit{Q1. Does the revised target align with the original query?}, we utilize a prompt in Figure~\ref{fig:filtering_c1_prompt}.
To select high relevant revised target for the given query, we measure scores for the given score rubric from 1 to 5 and corresponding explanation.
Average score distribution for this step is in Figure~\ref{fig:filtering_c1}.
When we randomly sample the outputs, score exceeding 3 out of 5 shows good quality.
After this process, 15,669 instances are survived out of 16,157 instances.
Detailed examples with both high and low scores are available in the Table~\ref{appendix:filtering_Q1}.
Next, for the second criterion, \textit{Q2. Does the revised target align with the given query and instructions, while other targets do not?.}, we use a prompt in Figure~\ref{fig:filtering_c23_prompt}.
In this step, we only select correct case that GPT-4 predicts annotated target among other distractors, and 11,992 instances are selected out of 16,157.
After merging these two steps, and select where more than 6 instances exist per group, we get total 1,267 groups with 9,906 instances left.

\begin{figure*}[t]
\begin{lstlisting}[style=data_prompt]
>>> SYSTEM PROMPT
You are a helpful, respectful and creative assistant.

>>> USER INSTRUCTION
You are a similarity evaluator! You're tasked with calculating the similarity between QUERY, and DOCUMENT displayed below based on their relevancy. In the evaluation, I want you to rate the relevancy of the pair according to the following score rubric:

Score 1: The DOCUMENT, QUERY have very little or no relevance to each other. The elements compared share almost no common attributes or context.
Score 2: The DOCUMENT, QUERY have some relevance but are quite distinct. They share a few attributes or contextual details, but there are significant differences in the majority of aspects.
Score 3: The DOCUMENT, QUERY are moderately relevant to each other. They share a fair amount of attributes or context, but there are still some notable differences that prevent a high similarity score.
Score 4: The DOCUMENT, QUERY have high relevance to each other. They share many attributes or contextual details, with only a few differences that do not majorly impact the overall similarity.
Score 5: The DOCUMENT, QUERY are very highly relevant or almost identical to each other. They share nearly all attributes or the context is almost exactly the same, with very minor or negligible differences.

You will be given QUERY and DOCUMENT pair.
[QUERY]
{{query}}

[DOCUMENT]
{{candidate document}}

You should generate based on the following format:

<Explanation>
{explanation for the score}
</Explanation>

<Score>
{score}
</Score>

Please give feedback on the DOCUMENT with respect to each QUERY, and provide a score on a scale of 1 to 5 whether it satisfies the requirements, where a higher score indicates better performance.
\end{lstlisting}
\caption{Prompt for filtering stage Q1 (step 4).}
\label{fig:filtering_c1_prompt}
\end{figure*}

\begin{table*}[t]
    \centering
    \small
    \setlength{\tabcolsep}{8pt}
    \begin{tabularx}{\textwidth}{>{\raggedright\arraybackslash}p{0.2\textwidth}|X}
        \toprule
        \textbf{Type} & \textbf{Example} \\
        \midrule
        Good & \textbf{[Query]} \\
        & Heteromorphic definition. \\
        & \textbf{[Target]} \\
        & In the realm of genetic research and study, "heteromorphic" specifically denotes the occurrence of distinct forms, or morphs, within an organism's life cycle, or among different organisms, as a result of genetic mutation. This term is often utilized in scholarly articles to describe variations that are not merely different in size but also morphologically distinct, such as varying chromosome shapes in the gametes of different sexes of the same species. In an educational setting, one would explore the significance of heteromorphism in understanding evolutionary processes and the resulting diversity among living forms. It's applied in studies that examine the genetic basis of these morphological variations and their implications for the survival and reproduction of species. \\
        & \textbf{[Score]} \\
        & 5 \\
        & \textbf{[Explanation]} \\
        & The document provides a detailed explanation of the term "heteromorphic" in the context of genetic research and study, explaining that it refers to the occurrence of distinct forms within an organism's life cycle or among different organisms as a result of genetic mutation. It also discusses the relevance of heteromorphism in evolutionary processes and the diversity of living forms. This explanation aligns well with the query, which asks for the definition of "heteromorphic." The document offers a clear, in-depth definition and contextualizes the term within its scientific application, demonstrating a high relevance to the query. \\
        \midrule
        Bad & \textbf{[Query]} \\
        & is catecholamine a steroid \\
        & \textbf{[Target]} \\
        & For those actively engaged in physical conditioning, it’s crucial to understand the role of various hormones. In the body's natural arsenal, there are hormones derived from the amino acid tyrosine that have significant roles, such as the thyroid hormone, which is pivotal for organ development and metabolic rate—an aspect vital for anyone with a focus on fitness. \\
        & \textbf{[Score]} \\
        & 1 \\
        & \textbf{[Explanation]} \\
        & the query is asking whether catecholamine is a steroid, which is a question about categorization of a substance based on its chemical nature. the document discusses hormones derived from the amino acid tyrosine, mentioning thyroid hormone and its role in metabolism and organ development, which is related to the broader topic of hormones. however, the document does not specifically address catecholamines or steroids, nor does it provide information to determine whether catecholamines are steroids. therefore, while there is a connection in the general subject matter (hormones), the specific question posed by the query is not answered by the document.\\
        \bottomrule
    \end{tabularx}
    \caption{Dataset example for filtering stage Q1 (step 4).}
    \label{appendix:filtering_Q1}
\end{table*}

\begin{figure*}[t]
\begin{lstlisting}[style=data_prompt]
>>> SYSTEM PROMPT
You are a helpful, respectful and creative assistant.

>>> USER INSTRUCTION
You are a ranker agent! Each potential DOCUMENT has a corresponding DOCUMENT id and you're tasked with ranking the answers based on their relevancy to the pair of QUERY, SCENARIO pair. In the evaluation, I want you to rate the relevancy of the pair according to the following score rubric:

Score 1: The DOCUMENT lacks relevance to the user's SCENARIO, providing little to no connection to the user's job, background, situation, location, occupation, hobbies, interests, or goals. It fails to consider preferences and context, resulting in an overall inadequate fit.
Score 2: The DOCUMENT has limited relevance, with only a few elements aligning with the user's SCENARIO and QUERY. While some contextual understanding and preference consideration may be present, it falls short of providing a comprehensive and well-fitted response.
Score 3: The DOCUMENT demonstrates moderate relevance, capturing some aspects of the user's SCENARIO. It shows an adequate contextual fit and considers a majority of the user's stated preferences. However, there is room for improvement in terms of depth and clarity.
Score 4: The DOCUMENT exhibits high relevance, aligning well with the user's SCENARIO and covering most relevant aspects. It demonstrates a strong contextual fit, addresses the user's preferences effectively, and maintains high clarity and conciseness. However, there may be minor areas for improvement.
Score 5: The DOCUMENT is perfectly relevant, precisely addressing all aspects of the user's SCENARIO, QUERY, and preferences. It seamlessly integrates with the user's context, demonstrating a profound understanding. The DOCUMENT is exceptionally clear, concise, and exhaustive in providing information, offering a flawless fit.

You SHOULD ONLY generate the top ranked id from the given search DOCUMENT (id : 1~10) and no additional comments as [id]. This is VERY IMPORTANT!

You will be given list of DOCUMENT and a pair of QUERY,SCENARIO.

[DOCUMENT LIST]
[1] {DOCUMENT_1}
[2] {DOCUMENT_2}
[3] {DOCUMENT_3}
...
[10] {DOCUMENT_10}

[SCENARIO]
{scenario}

[QUERY]
{query}

You should generate based on the following format:
<Explanation>
{explanation for the ranking}
</Explanation>

<Ranking>
{top ranked DOCUMENT id}
</Ranking>

Please give a top ranked DOCUMENT id with respect to each SCENARIO and QUERY pair, and provide a score on a scale of 1 to 5 whether it satisfies the requirements, where a higher score indicates better performance. 
\end{lstlisting}
\caption{Prompt for filtering stage Q2 (step 4).}
\label{fig:filtering_c23_prompt}
\end{figure*}

\section{Details for Human Verification}
\label{appendix:human_verification_details}

\begin{figure*}[t]
    \centering
    \subfloat[Instructions for Annotators.]{
        \centering
        \includegraphics[width=1.\textwidth]{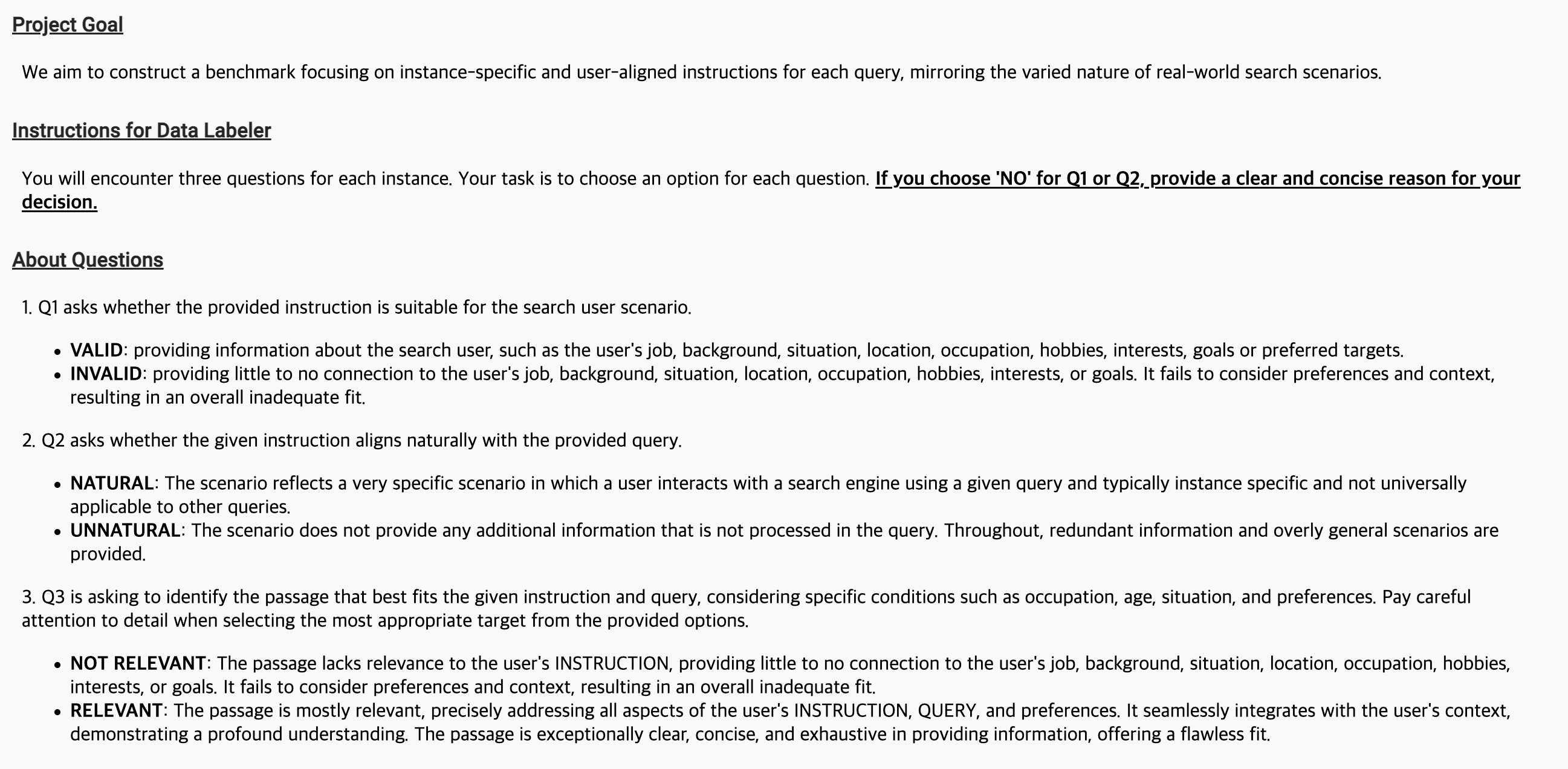}
    } \\
    \subfloat[Questions for each instance.]{
        \includegraphics[width=1.\textwidth]{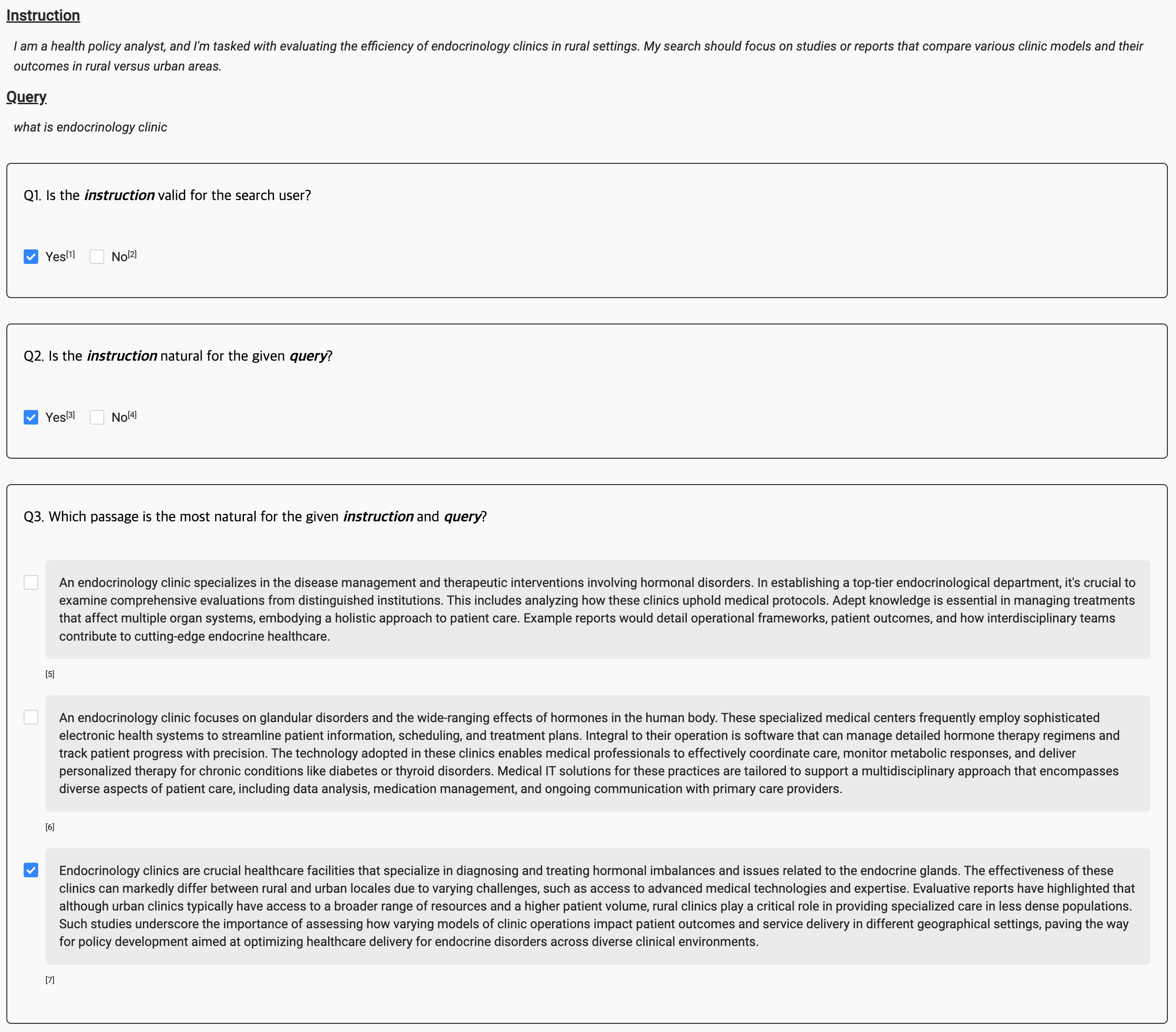}
    }
    \caption{User Interface for Human Verification.}
    \label{fig:human_verification_user_interface}
\end{figure*}

\paragraph{User Interface.}
Ten annotators are instructed to answer three questions as illusted in Figure~\ref{fig:human_verification_user_interface}. For the first question (\textit{Q1: Is the instruction valid for the search user?}) and second question (\textit{Q2: Is the instruction natural for the given query?}), annotators are required to choose either \textit{Yes} or \textit{No}. If the answer is \textit{No}, annotator are also required to provide a short reason. In the third question (\textit{Q3: Which passage is the most natural for the given instruction and query?}), annotators are provided three passages from the same group, including the correct answer.

\paragraph{Evaluation Settings.}
For each instance, the responses from three annotators are reported and regard the final human judgement by majority voting. In the first and second questions, we assess the reliability of the annotator's responses through kappa coefficient~\citep{Randolph2005FreeMarginalMK} with the proportion of \textit{Yes} responses. In the third question, we compare the final human judgement and \ours dataset through kappa coefficient~\citep{Cohen1960ACO} and report the top-1 accuracy.

\paragraph{Inter-agreement.}
Following the ~\citet{Landis1977TheMO}, the kappa coefficient for each question is interpreted as follow: \textit{poor agreement} (< 0); \textit{slight agreement} (0.01-0.20); \textit{fair agreement} (0.21–0.40); \textit{moderate agreement} (0.41–0.60); \textit{substantial agreement} (0.61–0.80); \textit{almost perfect agreement} (0.81–1.00). The first question shows \textit{almost perfect agreement}, and the second question shows \textit{moderate agreement} among the three annotators for each instance. And the third question shows \textit{substantial agreement} between the human judgement and \ours dataset.

\begin{table}[H]
\centering
\resizebox{1.0\columnwidth}{!}{
\begin{tabular}{lr}
\toprule
 \textit{Q1. Is the instruction valid for the search user?} & \\
\midrule
 \hspace{1.0em}kappa coefficient~\citep{Randolph2005FreeMarginalMK} & 0.9133 \\
 \hspace{1.0em}Proportion of \textit{Yes} response & 100\% \\
\toprule
 \textit{Q2. Is the instruction natural for the given query?} & \\
\midrule
 \hspace{1.0em}kappa coefficient~\citep{Randolph2005FreeMarginalMK} & 0.5267 \\
 \hspace{1.0em}Proportion of \textit{Yes} response & 97\% \\
\toprule
 \textit{Q3. Which passage is the most natural for the given instruction and query?} & \\
\midrule
 \hspace{1.0em}kappa coefficient~\citep{Cohen1960ACO} & 0.6468 \\
 \hspace{1.0em}Top-1 Accuracy & 76.5\% \\
\bottomrule
\end{tabular}
}
\caption{Human verification results for \ours.}
\label{tab:human_verification}
\end{table}

\begin{figure}[t!]
\begin{minipage}{\textwidth}
\small
\centering
\includegraphics[width=0.98\textwidth]{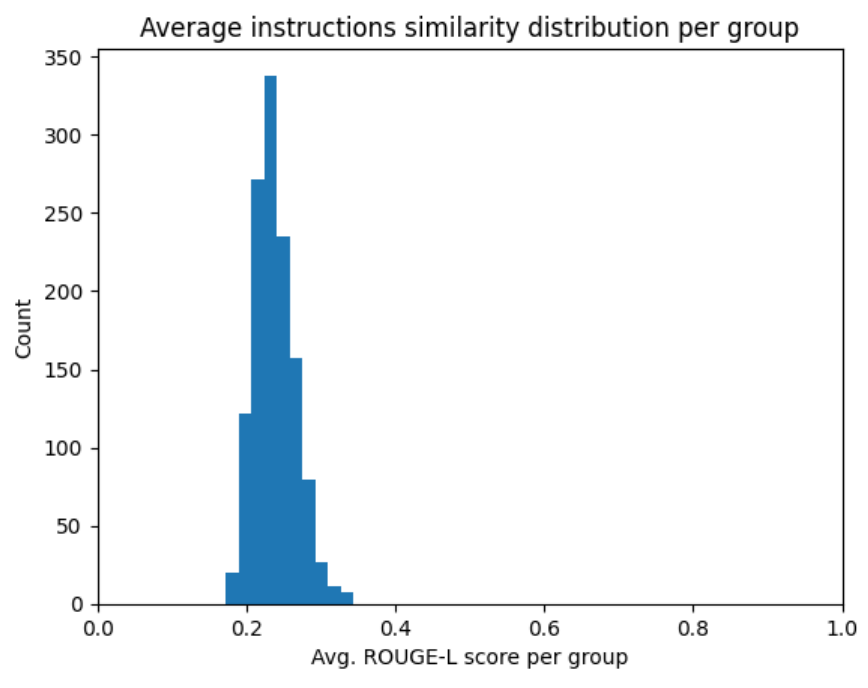}
\caption{ROUGE-L score distribution for the instructions.}
\label{fig:rouge}
\end{minipage}
\end{figure}

\begin{table*} [t]
\centering\
\resizebox{\textwidth}{!}{
\def\arraystretch{0.9}
\begin{tabularx}{\textwidth}{X}
\toprule
\small\textbf{[Query] \textit{what is the annual fee for spirit}} \\
\toprule
\small\textbf{[Instruction]} \\ 
\small I am a budget-conscious traveler looking to cut costs on my frequent trips. I prefer to invest more in experiences than in transport, so I need accurate information on the annual fee for Spirit Airlines' membership or credit card programs that offer benefits like waived bag fees or discounts. \\
\small\textbf{[Target]} \\ 
\small Thrifty explorers who habitually travel can find value in Spirit's credit card, especially since the \$59 fee is waived for the initial 12 months, allowing for savings early on. Although to offset the cost from the second year onwards, cardholders are expected to spend at least \$5,900 annually. Yet, the card can be lucrative for those who accumulate enough expenses, as it offers benefits like complimentary bag exemptions and fare reductions—perks that align with the preference to allot more for enriching experiences over transport expenses. \\
\midrule
\small\textbf{[Instruction]} \\ 
\small  As a travel agent, I'm preparing a cost-analysis presentation for a client who's interested in low-cost carriers. I am searching for the most up-to-date annual fee for Spirit's exclusive membership to include in my comparison chart alongside other budget airline fees. \\
\small\textbf{[Target]} \\ 
\small The exclusive membership offered by Spirit, recognized for its affordability, incurs an annual fee of \$59. This figure is essential for clients evaluating the comparative costs of low-cost airlines and determining the best value for their frequent travels. Notably, this fee is typically waived during the initial year of membership, enhancing the cost-effectiveness for new members. The fee is a crucial datapoint in the broader context of budget travel expenses and should be accounted for when mapping out annual travel budgets for cost-conscious clients. \\
\midrule
\small\textbf{[Instruction]} \\ 
\small  I am a college student on a tight budget with a penchant for exploring new places. I am researching various airlines' membership fees, particularly Spirit's annual fee, to figure out if the cost-saving benefits align with my limited resources and travel frequency. \\
\small\textbf{[Target]} \\ 
\small Balancing a wallet-friendly lifestyle while nurturing your wanderlust can be challenging, especially when it comes to airline memberships. Taking a closer look at Spirit's offerings, there's a \$59 yearly charge for their credit card, fortunately waived for the initial twelve months. To offset this expense, an expenditure of \$5,900 on the card annually is needed. One should consider their usual spending habits and the practicality of reaching such a sum in regular yearly outlays to determine if the membership aligns with prudent financial management as a diligent academic explorer. \\
\midrule
\small\textbf{[Instruction]} \\ 
\small I am a Human Resources professional planning the company’s travel budget for the upcoming fiscal year. I need to find the annual fee for Spirit to calculate whether bulk memberships for our staff travel would be advantageous compared to pay-per-use options. \\
\small\textbf{[Target]} \\ 
\small The annual membership fee for Spirit's corporate clients is designed to accommodate businesses looking to manage their travel expenses more effectively. While individual memberships or pay-per-use arrangements can potentially add to travel costs over time, opting for an annual corporate membership can offer substantial savings. This approach ensures that a company's travel budget is maximized, especially when considering the volume of staff travel. Keep in mind that corporate packages may offer additional perks, such as discounts on group bookings and other travel-related services, which further enhance the value proposition when compared to transaction-based fees. Evaluating these membership fees and aligning them with the company's projected travel volume will determine the most cost-efficient strategy for the fiscal year's travel budget. \\
\bottomrule
\end{tabularx}}
\caption{Dataset example in \ours benchmark.}
\label{appendix:dataset_example}
\end{table*}

\end{document}